\documentclass{article} 
\usepackage{nips15submit_e,times}
\usepackage{hyperref}
\usepackage{url}
\usepackage{amsmath}
\usepackage[lined]{algorithm2e}
\usepackage[pdftex]{graphicx}
\usepackage{subfigure}
\usepackage[hang]{caption}
\newtheorem{theorem}{Theorem}

\title{Causal Model Analysis using Collider v-structure with Negative Percentage Mapping}

\author{
Pramod Kumar Parida\\
Electrical \& Electronics Engineering Science \\
University of Johannesburg\\
South Africa \\
\texttt{pramodsstyle@gmail.com} \\
\And
Tshilidzi Marwala \\
Electrical \& Electronics Engineering Science\\
University of Johannesburg\\
South Africa \\
\texttt{tmarwala@gmail.com} \\
\And
Snehasish Chakraverty\\
Department of Mathematics \\
NIT Rourkela\\
India\\
\texttt{sne\_chak@yahoo.com} \\
}

\nipsfinalcopy 

\begin{document}

\maketitle

\begin{abstract}
	A major problem of causal inference is the arrangement of dependent nodes in a directed acyclic graph (DAG) with path coefficients and observed confounders. Path coefficients do not provide the units to measure the strength of information flowing from one node to the other. Here we proposed the method of causal structure learning using \textit{collider v-structures (CVS)} with \textit{Negative Percentage Mapping (NPM)} to get selective thresholds of information strength, to direct the edges and subjective confounders in a DAG. The NPM is used to scale the strength of information passed through nodes in units of percentage from interval $\left[0,1\right]$. The causal structures are constructed by \textit{bottom up} approach using path coefficients, causal directions and confounders, derived implementing collider v-structure and NPM. The method is self-sufficient to observe all the latent confounders present in the causal model and capable of detecting every responsible causal direction. The results are tested for simulated datasets of non-Gaussian distributions and compared with DirectLiNGAM and ICA-LiNGAM to check efficiency of the proposed method. 
\end{abstract}

\section{Background Study: Markov Model, d-separable, v-structures}\label{sec:1}

In Bayesian Learning the Markov properties play an important role, which set the criteria for structure analysis. The Markov model is a directed graph which does not contains any bi-directed edges [5]. This criteria does not provides the acyclicity of the model. But using markov model we can find the conditional dependent and independent nodes, provided we have information on prior observations. Sometime it happens to be very difficult to know about prior observations or any of the prior information. Assuming the observed variables are prior, the later ones will be conditionally dependent on the prior ones. Now we need a criteria to observe the posterior effects. In this scenario d-separable provides the criteria of sub-structural learning for conditional dependence and independence of the features. To implement d-separable we need v-structures.

The primary definition of v-structure is provided for the undirected graphical models, where two nodes are only connected through an undirected edge [3]. The v-structures in a markov model are the sub-structures where each three of the observed nodes are only connected through two directed edges. For the choice of any three nodes we can permute them in three possible ways using two directed edges irrespective of their positions. The d-separable criteria is used to construct a set of nodes such that conditioning on which, the paths/edges can be separated between other two nodes in the v-structure. Consider v-structures of $a \to b \to c$ , $ a \leftarrow b \to c$  and $a \to b \leftarrow c$. In first and second types $a$ and $c$ can be d-separated by observing/conditioning on $b$, while in the third one conditioning on $b$ makes $a$ and $c$ becomes dependent on each other as ancestor nodes of $b$ remains unobserved.. The third type of v-structure is called a collider.

He and Geng [2] applied intervention technique on markov equivalence classes to produce subgraphs with directed edges. They generated v-structures of subgraphs with markov properties to find directions, using conditional probabilities. Markov Blankets can reveal the directions in the nodes based on their relevance of connection inside the blanket. The structural learning of causal model can be shown through faithfulness and relevance of connections using d-separable and v-structures [6]. The faithfulness for markov equivalent classes are beneficial and weak or strong faithful criteria can be derived for Gaussian and uniform distributions, where path coefficients and errors are drawn from intervals $\left[-1,1\right]$ and $\left[0,1\right]$ respectively [9]. In following Sections \ref{sec:2} and \ref{sec:3}, we provided complete explanation for using collider v-structures as a sub–structural model and NPM for measuring strength of path coefficients. 
\section{Proposed Structural Learning Method}\label{sec:2}
Every orthodox and modern approach of causal construction lag to define the measure for information passing through directed edges, while path coefficient/connection strength is the information passed on. Depending on the information (path coefficient) either from nodes or external confounders we cannot conclude the directions in causal model. This problem becomes more complicated when it comes to multivariate structure. So we need a measure for these entropies through which we will be able conclude our directions.

 Before scaling the flow of information, we need a model to work on for causal constructions. The easiest way to build a larger model is, first to build the sub-structures of the bigger model and then arrange these small ones to complete the required structure. The directed acyclic causal graph can be studied as a decomposition of multiple v-structures under the lights of markov model [10]. In our case we used a \textit{bottom up} approach of arranging multiple v-structures to form a causal DAG. For this type of approach we don't need any prior information regarding the structure or contributing variables. After deriving all the direction, path coefficients and confounders for the considered v-structures, these can be combined together to provide a complete causal structure. The model of collider v-structure provides the necessary sub-structural analysis in a very detailed way.
 
\subsection{Collider v-structure with NPM (CVS with NPM):}

\begin{figure}[h]
\centering
\subfigure[]{%
	\includegraphics[width=0.25\linewidth, height=1in] {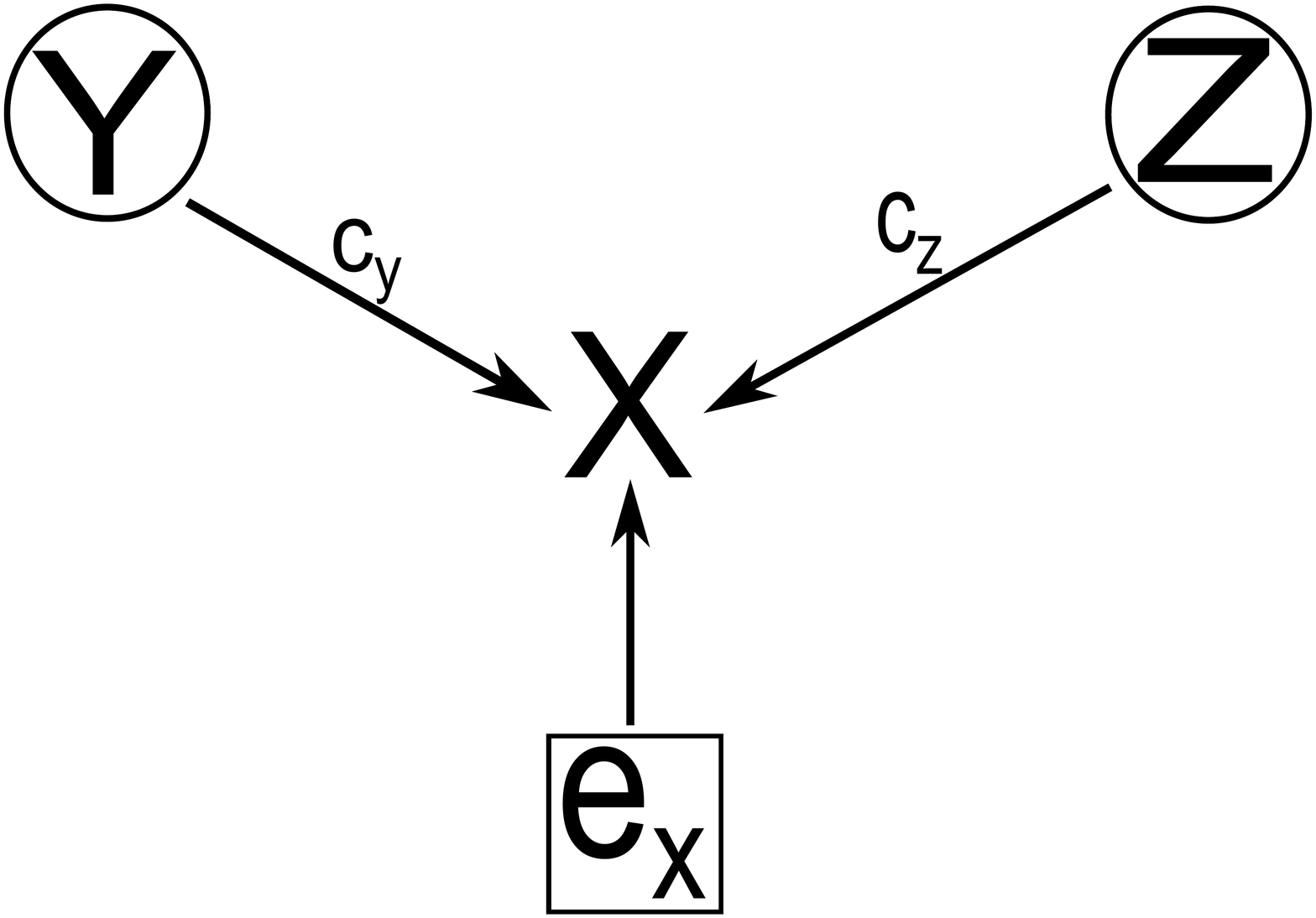}
	\label{fig:1a}}	
\quad
\subfigure[]{%
	\includegraphics[width=0.25\linewidth, height=1in] {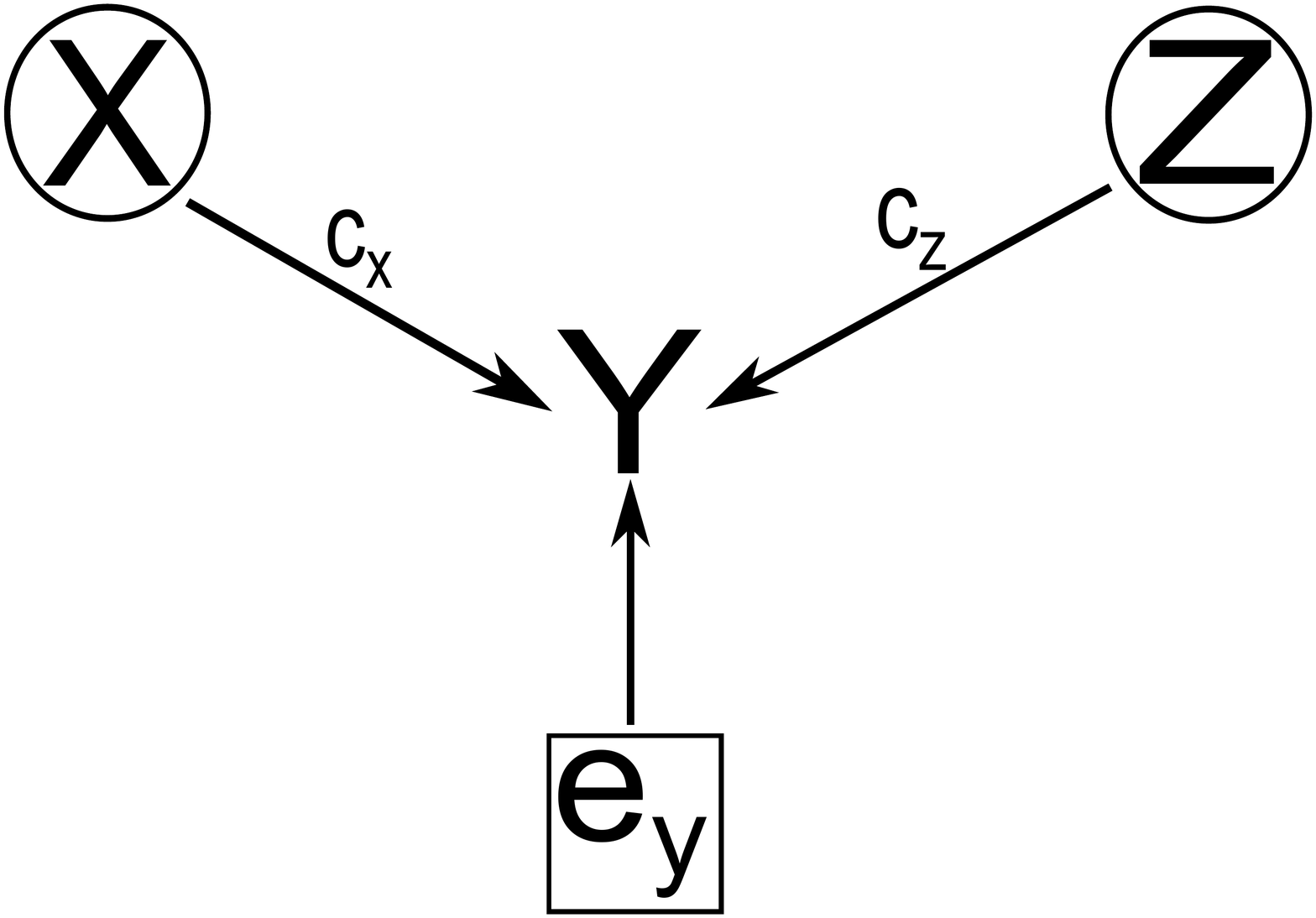}
	\label{fig:1b}}	
\quad
\subfigure[]{%
	\includegraphics[width=0.25\linewidth, height=1in] {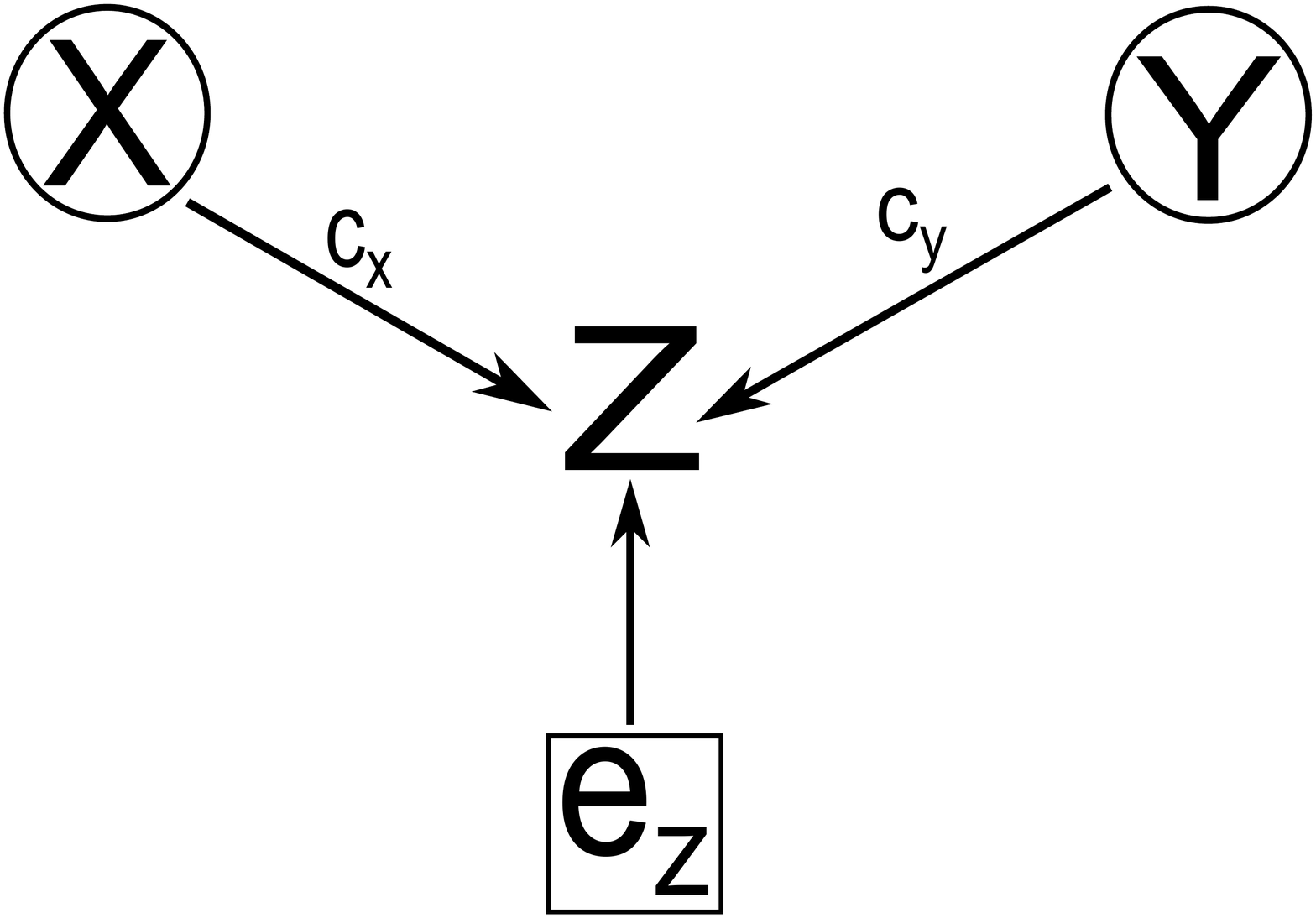}
	\label{fig:1c}}	
\caption{Collider v-structure}
\label{fig:1}
\end{figure}

To analyze a larger causal structure in a smaller sub-structural level consider the v-structure as defied in the markov model, that is a set of three nodes connected through two directed edges. The collider is a v-structure where two extreme nodes have directed edges towards the middle node. We referred it as \textit{collider v-structure} whenever an external confounder is present or added to the collider system and we modeled our algorithm suitable to analyze these types of DAG. Below Figure \ref{fig:1} provides the collider v-structures. 

Consider the collider v-structures shown in Figure \ref{fig:1}(a). A basic collider system where $Y$ and $Z$ collides with $X$. We referred the extreme nodes $Y$ and $Z$ as colliders and $X$ as collide.  The values $c_y, c_z$ are path coefficients of $Y, z$ respectively and $e_x$ is the error/confounder added to node $X$. Figure \ref{fig:1}(b) and \ref{fig:1}(c) are the permutations of nodes shown in Figure \ref{fig:1}(a). Now Figure \ref{fig:1}(a) can be represented as $X = (c_y)Y + (c_z)Z + e_x$ which is a linear equation in multivariate case. Now following points provide better explanation for choice of collide v-structure,

\begin{itemize}

\item Collider v-structures are small and easy to infer. The complete inference technique is described in details latter in the Section \ref{sec:3}.

\item In collider state the two extreme nodes become conditionally dependent whenever we observed or conditioned the middle node. So if we will compute the middle node as the combination of two extreme nodes and additive noise (confounder), then the collider v-structure is itself becomes a causal model which holds all properties of a complete causal structure.

\item Inference using successive permutations on selected nodes in the collider v-structure provides easier access to conditional dependence or independence of all three nodes. Solving for coefficients of collider nodes and confounder, the estimations for directions can be made using a threshold criteria.

\end{itemize}

To direct the edges in a collider v-structure based on their contribution through path coefficients, we introduced the negative percentage mapping (NPM) technique. NPM provides the percentage contributions of path coefficients of collider nodes and confounders in the collider v-structure towards the middle node (collide). In Figure \ref{fig:1}(c), we know the path coefficients for $X$ and $Y$ towards $Z$. But we don’t know the percentage contributions of ${c_x}X$, ${c_y}Y$ and $e_z$ towards $Z$, that is how strongly these values are contributed toward the formation of $Z$. Let's dig a bit more to understand what percentage contribution is. Consider a linear model $c=a(x)+b$  given by a simple numeric representation of $10=2.37(5.45)-2.92$. Now if we compute $10$ as total of $100\%$, then percentage contribution of ($2.37(5.45), -2.92$) are ($129.2\%, -29.2\%$) respectively. The percentage contributions above $100\%$ are referred as over margin percentage and negative contributions are referred as negative percentages. Now the problem arises with $-29.2\%$ which either mean that $-2.92$ has no contribution towards $10$ or contributes as negative percentage. Next question is how contribution becomes greater than the total estimation itself as in case of $2.37(5.45)$. In real world the negative and over margin contribution does not make any sense as they must have some contribution towards the formation of total percentage. This is where we need the NPM technique to map negative and over margin percentages to some form of positive contribution under the margin of $100\%$. Following points are discussed why NPM is required,
\begin{itemize}

\item Negative percentage mapping provides a way to map the negative percentages and over margin percentage contributed towards the margin of $100\%$.

\item Using NPM, we can optimize the path coefficients of collider nodes to find which one provides the best estimation of direction for the choice of collider v-structure.

\item The higher NPM values confirm strong existence of directions in the v-structure. In computational process of NPM we maximized for the effect of path coefficients on collide node. Therefore the directions produced in the nodes provide best estimations.

\end{itemize}

Now the question is why we need to maximize percentage contributions for path coefficients and are we minimizing the contributions of confounders. In collider v-structure our main goal is to find the directions in nodes, irrespective of the larger or smaller contributions of errors. We constructed our model such a way that maximization of path coefficient do not implies the necessary minimization for errors/confounders. 

In Figure \ref{fig:1}, (a), (b) and (c) are permutations of nodes $X, Y$ and $Z$. So why we need the permutations over nodes to find the directions in $X, Y$ and $Z$. Let's see the following points,

\begin{enumerate}
	
	\item In Figure \ref{fig:1} (c), $X$ and $Y$ have some fixed contribution towards $Z$ irrespective of their coefficients. Let the contributions be $x_p$ and $y_p$ for $X$ and $Y$ respectively.
	
	\item The coefficients $c_x$ and $c_y$ of $X$ and $Y$, either increase or deceases the contributions $x_p$ of $X$ and $y_p$ of $Y$ towards $Z$. This increment or decrement depend on factors like error and others nodes. So contribution of $c_xX$ depends on contribution of $c_yY$ and $e_z$ toward the formation of $Z$.
	
	\item Using above criteria, we can observe how the contribution between two nodes can be maximized under the influence of other nodes. This can be done by checking the contributions in two particular nodes towards each other at presence of other nodes. This process is called the permutation.
\end{enumerate}

In the next Section we provided a complete mathematical model for collider v-structure and NPM for experimental use.
\section{Mathematical model: Collider v-structure, NPM}\label{sec:3}

To provide a mathematical model, we need to simplify the causal structures shown in Figure \ref{fig:1}. We also need a generalization for the possible permutations in the collider v-structure.

\subsection{Model for Collider v-structure:}

Consider a set of observations\(V\)of\(m \times n\) matrix, where \(m\) is the number of instances and \(n\) is the number of variables or observations. Now draw any three set of observations \((i,j,k)\) to construct the collider v-structure with fully observed confounders. Below equation is a generalized form of linear model of collider v-structure,

\begin{equation}\label{eq:1}
\sum\limits_{i = 1}^m {{v_i}}  = {c_{ji}}\sum\limits_{j = 1}^m {{v_j}}  + {c_{ki}}\sum\limits_{k = 1}^m {{v_k}}  + {e_i}
\end{equation}	
					
In equation \ref{eq:1}, \({v_i}\) is the middle collide and \({v_j},{v_k}\) are the extreme colliders, \(({c_{ji}},{c_{ki}})\) represents path coefficients of \(({v_j},{v_k})\) respectively while the subscripts \((ji,ki)\) in coefficients represents the directions from \((j \to i,k \to i)\) respectively and \({e_i}\) is the confounder added to \({v_i}\).  Now the required permutations of nodes in equation \ref{eq:1} can be given as,

\begin{equation}\label{eq:2}
\begin{array}{l}
\sum\limits_i {{v_i}}  = {c_{ji}}\sum\limits_j {{v_j}}  + {c_{ki}}\sum\limits_k {{v_k}}  + {e_i},\\
 \sum\limits_j {{v_j}}  = {c_{ij}}\sum\limits_i {{v_i}}  + {c_{kj}}\sum\limits_k {{v_k}}  + {e_j},\\
 \sum\limits_k {{v_k}}  = {c_{ik}}\sum\limits_i {{v_i}}  + {c_{jk}}\sum\limits_j {{v_j}}  +{e_k}
\end{array}
\end{equation} 					

Equation \ref{eq:2} can be solved using multivariate least square regression technique by taking derivative over required parameters and equating them to zero. In equation \ref{eq:2}, the parameters are  path coefficients and errors. Solving for three linear models given in equation \ref{eq:2}, we can find the following matrices,

\begin{equation}\label{eq:3}
\begin{array}{l}
\left[{\begin{array}{*{20}{c}}
	{\sum {{v_j}{v_j}} }&{\sum {{v_j}{v_k}} }&{\sum {{v_j}} }\\
	{\sum {{v_k}{v_j}} }&{\sum {{v_k}{v_k}} }&{\sum {{v_k}} }\\
	{\sum {{v_j}} }&{\sum {{v_k}} }&1\end{array}} \right]
\left[{\begin{array}{*{20}{c}}{{c_{ji}}}\\{{c_{ki}}}\\{{e_i}}\end{array}} \right] = 
\left[{\begin{array}{*{20}{c}}{\sum {{v_i}{v_j}} }\\
	{\sum {{v_i}{v_k}} }\\
	{\sum {{v_i}} }\end{array}} \right]...(i)\\

\left[{\begin{array}{*{20}{c}}
	{\sum {{v_i}{v_i}} }&{\sum {{v_k}{v_i}} }&{\sum {{v_i}} }\\
	{\sum {{v_i}{v_k}} }&{\sum {{v_k}{v_k}} }&{\sum {{v_k}} }\\
	{\sum {{v_i}} }&{\sum {{v_k}} }&1\end{array}} \right]
\left[{\begin{array}{*{20}{c}}{{c_{ij}}}\\{{c_{kj}}}\\{{e_j}}\end{array}} \right] = 
\left[{\begin{array}{*{20}{c}}
	{\sum {{v_j}{v_i}} }\\
	{\sum {{v_j}{v_k}} }\\
	{\sum {{v_j}} }\end{array}} \right]...(ii)\\

\left[{\begin{array}{*{20}{c}}
	{\sum {{v_i}{v_i}} }&{\sum {{v_j}{v_i}} }&{\sum {{v_i}} }\\
	{\sum {{v_i}{v_j}} }&{\sum {{v_j}{v_j}} }&{\sum {{v_j}} }\\
	{\sum {{v_i}} }&{\sum {{v_j}} }&1\end{array}} \right]
\left[{\begin{array}{*{20}{c}}{{c_{ik}}}\\{{c_{jk}}}\\{{e_k}}\end{array}} \right] = 
\left[{\begin{array}{*{20}{c}}{\sum {{v_k}{v_i}} }\\
	{\sum {{v_k}{v_j}} }\\
	{\sum {{v_k}} }\end{array}} \right]...(iii)
\end{array}	
\end{equation}
		
In equation \ref{eq:3}, matrices shown in (i), (ii) and (iii) provides parameters of equations \ref{eq:2} when solved for variables \({v_i},{v_j}\) and \({v_k}\) respectively.  The results of equation \ref{eq:3} can be written as

\begin{equation}\label{eq:4}
\left[ {\begin{array}{*{20}{c}}0&{{c_{ji}}}&{{c_{ki}}}\\{{c_{ij}}}&0&{{c_{kj}}}\\
	{{c_{ik}}}&{{c_{jk}}}&0\end{array}\begin{array}{*{20}{c}}{{e_i}}\\{{e_j}}\\{{e_k}}\end{array}}\right]
\left[ {\begin{array}{*{20}{c}}{{v_i}}\\{{v_j}}\\{{v_k}}\\1\end{array}} \right] = \left[ {\begin{array}{*{20}{c}}{{v_i}}\\{{v_j}}\\{{v_k}}\end{array}} \right]
\end{equation}
						
Now if any of the path coefficients in equation \ref{eq:4} are zero then we can directly conclude that there exist no direction in the observed nodes. But, if all the estimated parameters in matrix are nonzero, then the problem arises for selecting the best parameters estimated for path coefficients producing directions in nodes and the confounders. So we need a threshold criteria for parameter selection, above which we can conclude the directions in nodes. This threshold is computed as the percentage contributions of colliders and confounders toward the collide node. As discussed in Section \ref{sec:2}, the general computation for percentage contribution does not provide scope for negative and over margin percentages. In the following paragraphs we provided the proposed method for negative percentage mapping (NPM). 

\subsection{Negative Percentage Mapping (NPM):}

To compute the thresholds for selection of best possible directions in the collider v-structure, we need to find the percentage contribution for each of the observed path coefficients and confounders estimated in equation \ref{eq:3}.

\begin{theorem}\label{thm:1}
Consider the linear model for two variable set \((x,y) \in \Re \) as \(y = ax + b\), where \(y\) estimated as a function of \(x\) and \(a,b\) are estimation parameters to fit the liner model. The percentage contribution (\({\% ^c}\)) for \((ax,b) \in \Re \) is $ (ax,b){\% ^c} = {(ax,b)/y} $ always holds, for each $\{ (ax){\% ^c},(b){\% ^c}\}  \in {[0,1]} $ such that \((ax){\% ^c} + (b){\% ^c} = 1\). If \((ax){\% ^c} + (b){\% ^c} = 1\) for any \(\{ (ax){\% ^c},(b){\% ^c}\}  \in S\), where set \(S\) contains the elements from \(\{ [-\infty,-1], [-1,0], [0,1], [1,\infty]\} \) then for every over margin percentage there exist a negative percentage for wich \((ax){\% ^c} + (b){\% ^c} = 1\) holds. These percentages can be mapped to interval \([0,1]\) by negative percentage mapping defined as \[{(ax,b){^n\%^c}} = {\{|{(ax)}{\%^c}|,|{(b){\%^c}}|\}}/{\{|{(ax)}{\%^c}| + |{(b){\%^c}}|\}}.\] 		
\end{theorem}

The above idea can be extended to a multivariate vector space. Now why it's called a negative percentage mapping? Consider the case of three variable linear model defined in equation \ref{eq:1}. The parameters \(({c_{ji}},{c_{ki}},{e_i})\) can take any values from set \(S\), where each parameter can attain any value from 4 intervals sets of \(S\). An observable fact is that over margin percentage implies existance negative percentage irrespective of their signs (+,-). So if we will observe all the negative percentage contributions, then through negative percentage mapping we can easily compute all the parameter contributions in \([0,1]\) interval. So we called it a negative percentage mapping.

Using theorem \ref{thm:1} in equation \ref{eq:4}, the conclusions are drawn on path coefficients and confounder responsible for causal directions. Threshold criteria for selection of directions are discussed in following points.

\begin{enumerate}

\item Find the \({\% ^c}\) for parameters in equation \ref{eq:4}. Compare \({\% ^c}\) of \(({c_{ij}},{c_{ji}}),({c_{ik}},{c_{ki}}),({c_{jk}},{c_{kj}})\) and perform NPM. Select highest percentage for the observed coefficient sets.

\item In equation \ref{eq:2}, it is not possible to have\(({c_{ij}}{\% ^c} = {c_{ji}}\% ),({c_{ik}}{\% ^c} = {c_{ki}}{\% ^c}),({c_{jk}}{\% ^c} = {c_{kj}}{\% ^c})\). If it happens then discard the values as it create cycles in causal structure.
\end{enumerate}

Using the collider v-structure and NPM, a complete causal DAG is fully traceable with completely observed set of confounders. In next Section we discussed the experimental set ups for randomizes experiment and implementable algorithm.

\section{Experimental Setup}\label{sec:4}

For the best machine implementation of our proposed method an optimized and fully randomized experimental setup is used. The optimized algorithm for causal inference using collide v-structure and NPM takes an matrix desired for causal inference and provides four matrices of strength, direction, error matrix and the percentage contribution matrix. For algorithm design following assumptions are made,

\begin{enumerate}

	\item Input is a matrix \(V\) of \(m \times n\) elements, where \(m\) is the number of successive observations and \(n\) is the number of variables.
	
	\item	Construct a matrix \(C\), where each column of \(C\) contains a randomized set of 3 combinations from \(n\) variables. Constructing collider v-structure for all sets of nodal and confounder combination, we certainly make assured that there exist no unobserved confounder and causal directions in the nodes.	
	
	\item	Set the threshold to the highest \({\% ^c}\) value observed for a set of nodes in current state when compared to previously observed \({\% ^c}\) values. Strength, direction, error and \({\% ^c}\) matrices are successively updated based on threshold condition.
	
	\item	Output matrices of strength and error contain the path coefficients and confounder values for which \({\% ^c}\) are highest. Direction matrix saves a count for directions in the nodes observed through threshold condition and the matrix \({\% ^c}\) records the highest \({\% ^c}\) values responsible for directions in any two sets of nodes, whenever threshold values get updated. The \(i,j\)  value for strength and error shows the strength from node $i$ to $j$ and  in error matrix the column $j$ shows the cumulative error sums at node $j$.  

\end{enumerate}
A complete Implementable algorithm is given below for causal structure learning using CVS with NPM.
\begin{algorithm}[h]
	\KwIn{A matrix \(V\) of size \(m\times n\)}
	\KwOut{\(STRN, DRCT, PCNT\) matrices each of size \(n \times n\) and \(ERR\) is a matrix of size \(1 \times n\)}
	\caption{Causal structure learning using collider v-structure and NPM}
	\label{alg:1}
	\BlankLine
	Construct matrix \(C\) of size \(3 \times {n \choose 3}\) \;
	Initialize all output matrices to Zero \;		
	\For{\(i \leftarrow 1\) \KwTo combinations in \(C\)}{
		Construct \(R\) by randomly selecting colunms from \(C\) \;
		Solve for Parameter sets \((c_{ji},c_{ki},e_i),(c_{ij},c_{kj},e_j),(c_{ik},c_{jk},e_k)\) \;
		Store  path coefficients in the matrix $PARAM$ and errors in \(ERS\) \;
		Compute the \(\%^c\) of matrix \(PARAM, ERS\) \;			
		\If{ any \(\%^c\) of (\(PARAM or ERS < 0\)) }{
				Compute \(^n\%^c\) matirx by applying \(NPM\) on \(PARAM \& ERS\)					
		}
		\For{ \(p,q\): rows \& columns in \(R\) }{
			\If{\(^n\%^c\) of \(PARAM[p,q] > PCNT[p,q]\)}{
				Set a count for directions and Store it in \(DRCT\) \;
				Update \(STRN, ERR \& PCNT\) using \(PARAM, ERS \& ^n\%^c\)\;
			}	
		}
		\For {\((r,c)\) in rows \& columns of \(PCNT\)}{
			\If{\(PCNT[r,c] > PCNT[c,r]\)}{
				\([c,r]\) of \(STRN, DRCT, ERR, PCNT = 0\) \;
			}				
		}				
		\Return {\(STRN, DRCT, ERR, PCNT\)}\; 		
	}				
\end{algorithm}

The Experiment is designed and carried out in R environment. The Computational complexity of the algorithm is \(O{({nC3} + n^2) }\). Which makes this method faster than other methods having a computational complexity of \({O(n^3)}\). Due to short computation time this method can be used to analyze large datasets.

\section{Simulations}\label{sec: 5} 

For simulation we produced synthetic datasets to check the performance of our proposed model. The proposed model is duly tested for data of type Gaussian, non-Gaussian and noisy. Bach and Jordan [1] provides the framework to construct the sparse matrices with known probabilities. The non-Gaussianity of data is more useful than Gaussian for identification of causal DAGS [7]. In our case data is synthesized using following properties

\begin{enumerate}
	\item Construct 4 dataset with \(m=500,1500,2000,3000\) number of instances and \(n=10,25,35,45\) number of variables.
	
	\item For each \(n\) number of variable set, draw $40\%$ of i.i.d samples from Gaussian and non-Gaussian distributions of the types (1) Uniform, (2) Exponential, (3) Normal, (4) Log-normal, (5) Laplace, (6) Student- t with 3 degree freedom, (7) Student –t with 5 degree freedom, (8)-(28) Symmetric mixtures of any two types from (1) to (7), (29)-(49) Non-symmetric mixture of any two types from (1) to (7).
	
	\item Now remaining $60\%$ of the samples for each \(n\) variables is drawn from random mixing of different distributions from (1) to (49) in a linear model as defined in equation \ref{eq:1}. This results in a noisy dataset with correlated error.
	
	\item For linear mixture model, the mixing coefficients are randomly drawn from intervals \(\left[ {- 5,- 0.5} \right] \cup \left[{0.5,3} \right]\) and confounders are drawn from interval \(\left[{- 2,3} \right]\). 
\end{enumerate}

The model is tested for effective computational time with ICA-LiNGAM [7] and DirectLiNGAM [8] methods. The system used for simulation process is a dual core i3 variant. In comparison test the choice of environments are different (environments of ICA-LiNGM, DirectLiNGAM is MATLAB and CVS with NPM is R), so true efficiency is not available. Table \ref{tab:1} provides the results on time consumed by methods for different number of sample sets.
\begin{table}[H]	
	\captionsetup{width=0.7\textwidth}
	\caption{Computational times of CVS with NPM, ICA-LiNGAM and DirectLiNGAM methods for different i.i.d sample sets}		
	\label{tab:1}	
	\begin{center}		
		\begin{tabular}{|cc|cccc|}
			\hline
			Methods & Samples  & 500 & 1500 & 2000 & 3000\\ \hline
			CVS with NPM & 10 & 0.26 s & 0.31 s & 0.33 s & 0.34 s\\
			& 25 & 4.34 s & 4.34 s & 4.38 s & 4.47 s\\
			& 35 & 12.49 s & 12.66 s & 13.06 s & 13.14 s\\
			& 45 & 26.64 s & 27.03 s & 26.35 s & 26.62 s\\ \hline
			ICA-LiNGAM & 10	& 0.62 s &	0.64 s & 0.66 s	& 0.69 s\\
			& 25 & 1.98 s &	2.27 s & 2.68 s & 3.94 s\\
			& 35 &	4.18 s & 4.52 s & 4.65 s &	4.499 s\\
			& 45 &	5.27 s & 7.55 s & 7.23 s & 10.35 s\\ \hline
			DirectLiNGAM & 10 & 1.03 min & 9.49 min & 14.27 min & 25.40 min\\ 
			\hline	
		\end{tabular}			
	\end{center}	
\end{table}

\begin{figure}[H]
	\centering
	\subfigure[For 5 nodes]{%
		\includegraphics[width=0.4\linewidth, height=1in] {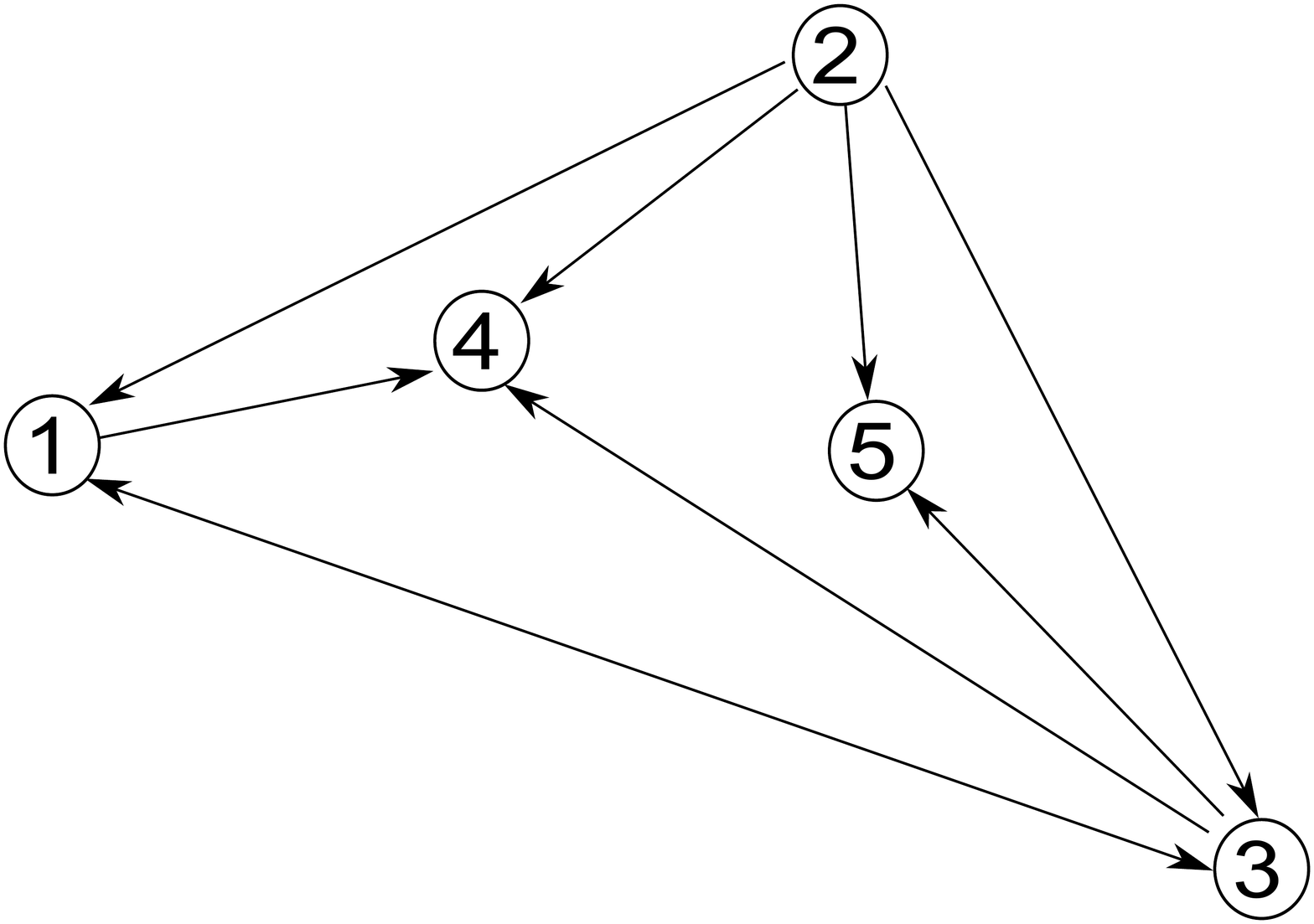}
		\label{fig:3a}}
	\quad
	\subfigure[For 8 nodes]{%
		\includegraphics[width=0.4\linewidth, height=1in] {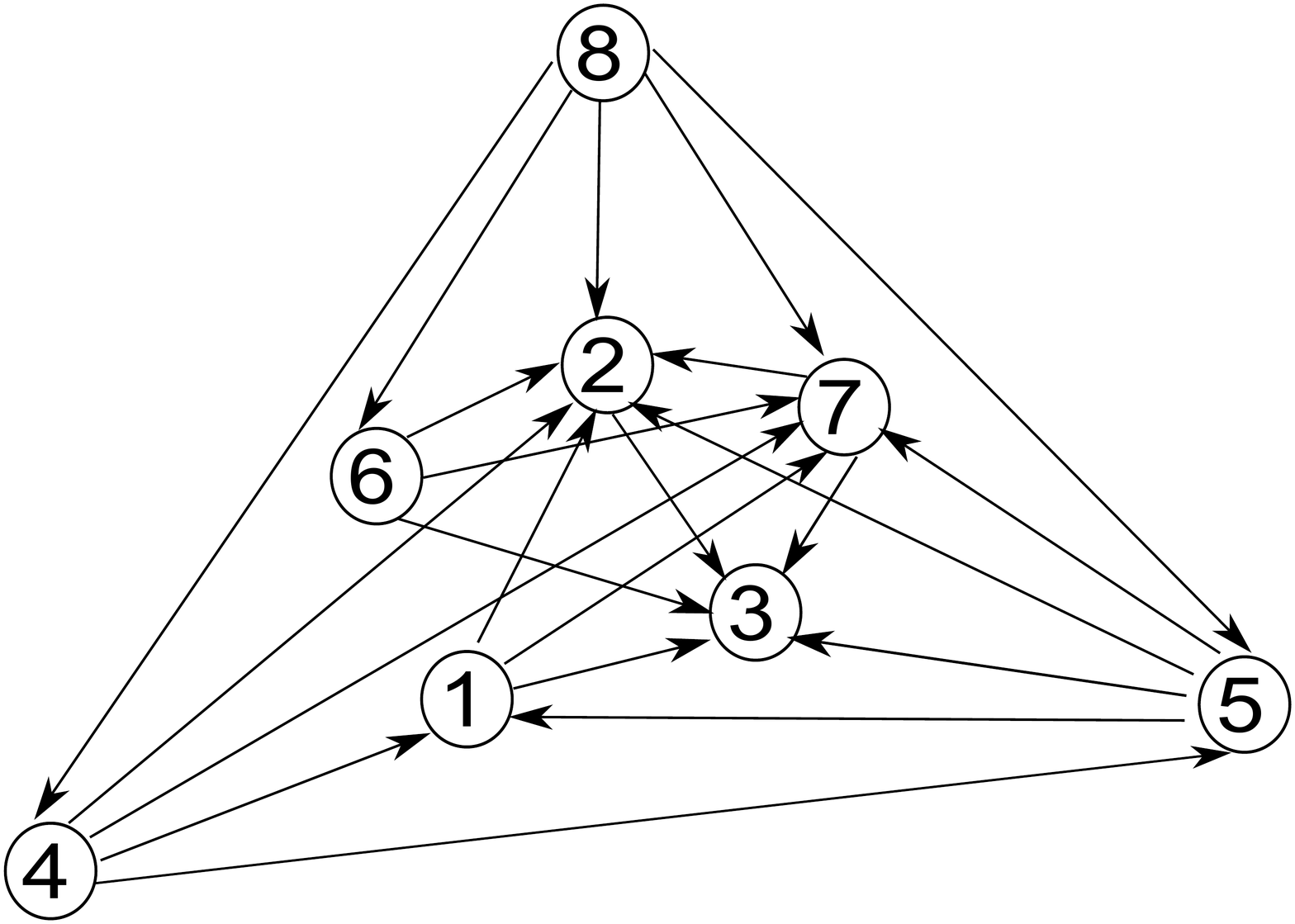}
		\label{fig:3b}}			
	\caption{Causal models using DirectLiNGAM for 5 and 8 no of nodes}
	\label{fig:3}
\end{figure}

\begin{figure}[H]
	\centering
	\subfigure[Using CVS with NPM]{%
		\includegraphics[width=0.4\linewidth, height=1in] {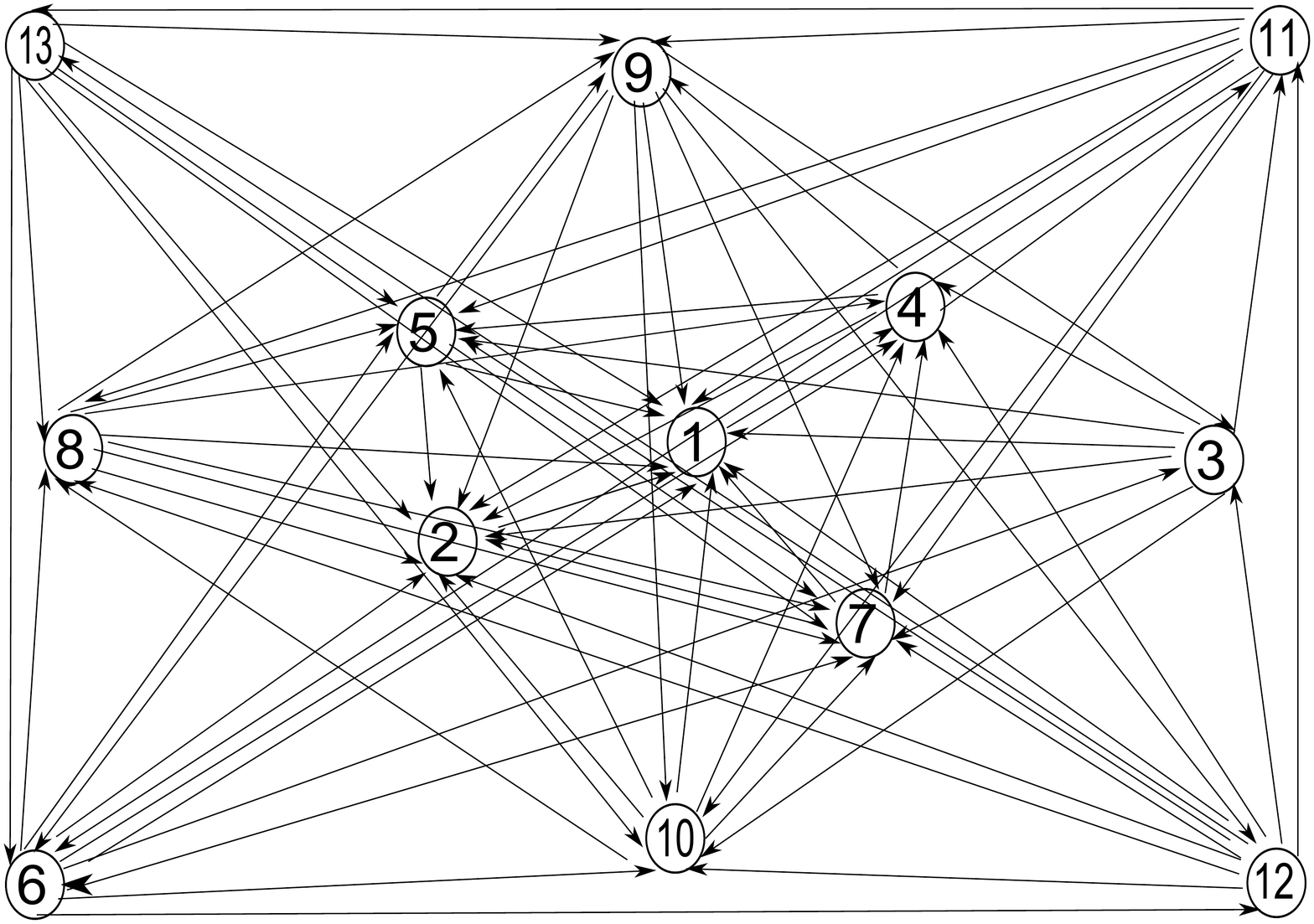}
		\label{fig:4a}}
	\quad
	\subfigure[Using DirectLiNGAM]{%
		\includegraphics[width=0.4\linewidth, height=1in] {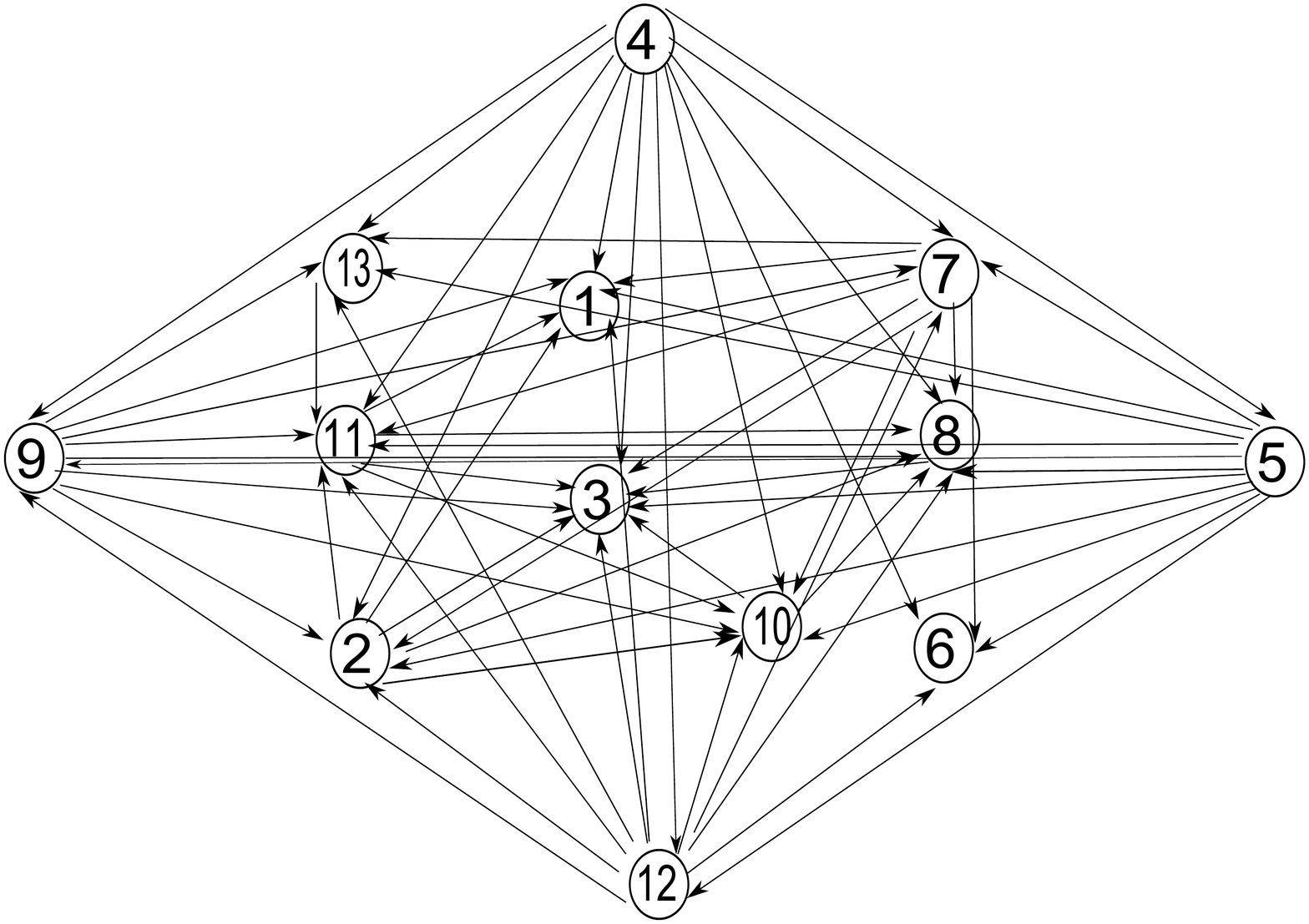}
		\label{fig:4b}}			
	\caption{Causal structured for Boston Housing Dataset using CVS with NPM and DirectLiNGAM}
	\label{fig:4}
\end{figure}

In our simulation test ICA-LiNGAM and DirectLiNGAM methods failed for noisy dataset, drawn from linear mixture model. The datasets compared in this simulation are i.i.d samples drawn from non-Gaussian distribution types from (1) to (49). For smaller datasets CVS with NPM method out performed ICA-LiNGAM, but ICA-LiNGAM performed better for larger sets. In all test sets CVS with NPM over shadowed DirectLiNGAM from the begining, so we did not provided the rest of the results for DirectLiNGAN. 
\begin{figure}[H]
	\centering
	\subfigure[For 5 nodes]{%
		\includegraphics[width=0.4\linewidth, height=0.75in] {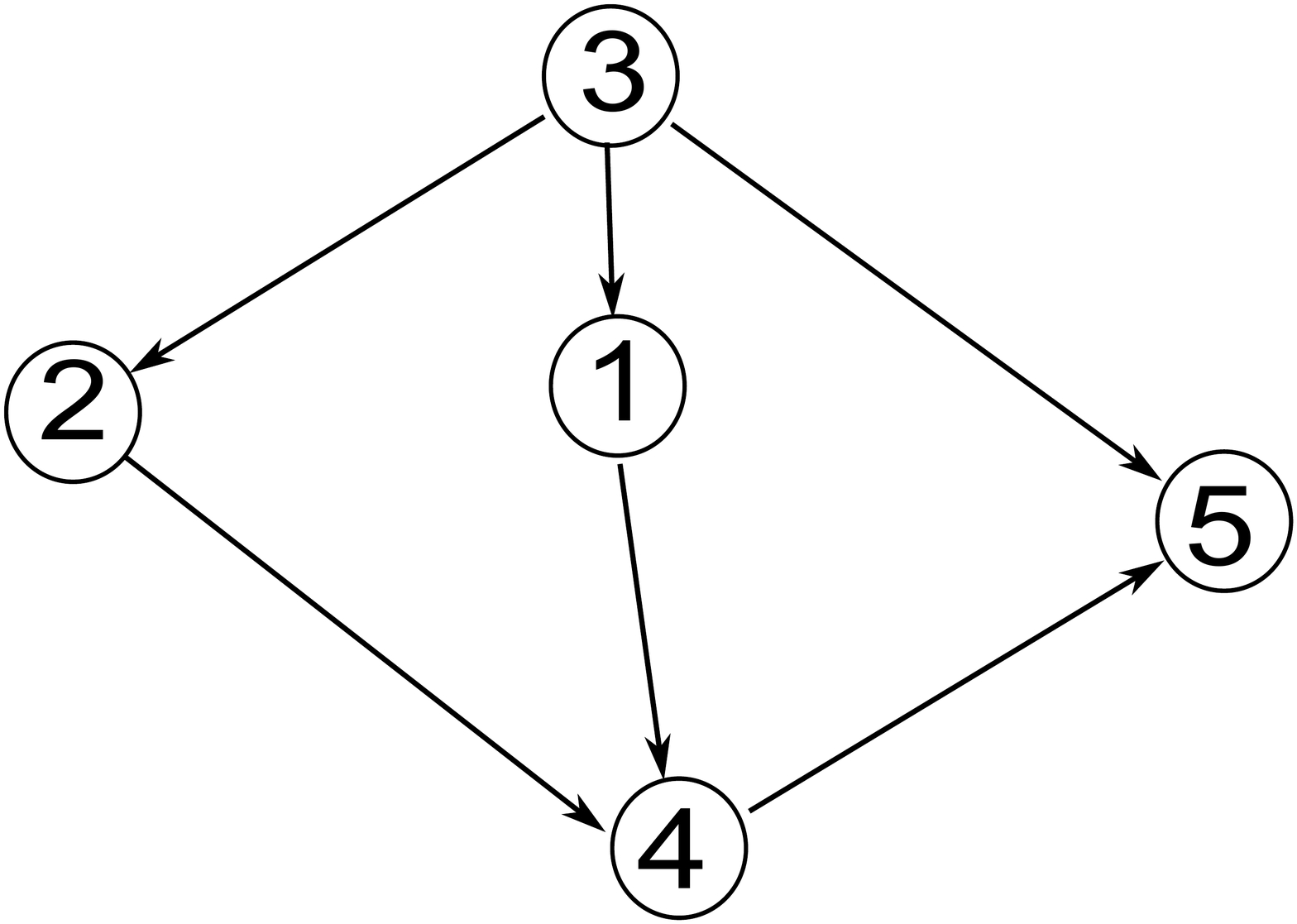}
		\label{fig:2a}}	
	\quad
	\subfigure[For 8 nodes]{%
		\includegraphics[width=0.4\linewidth, height=0.75in] {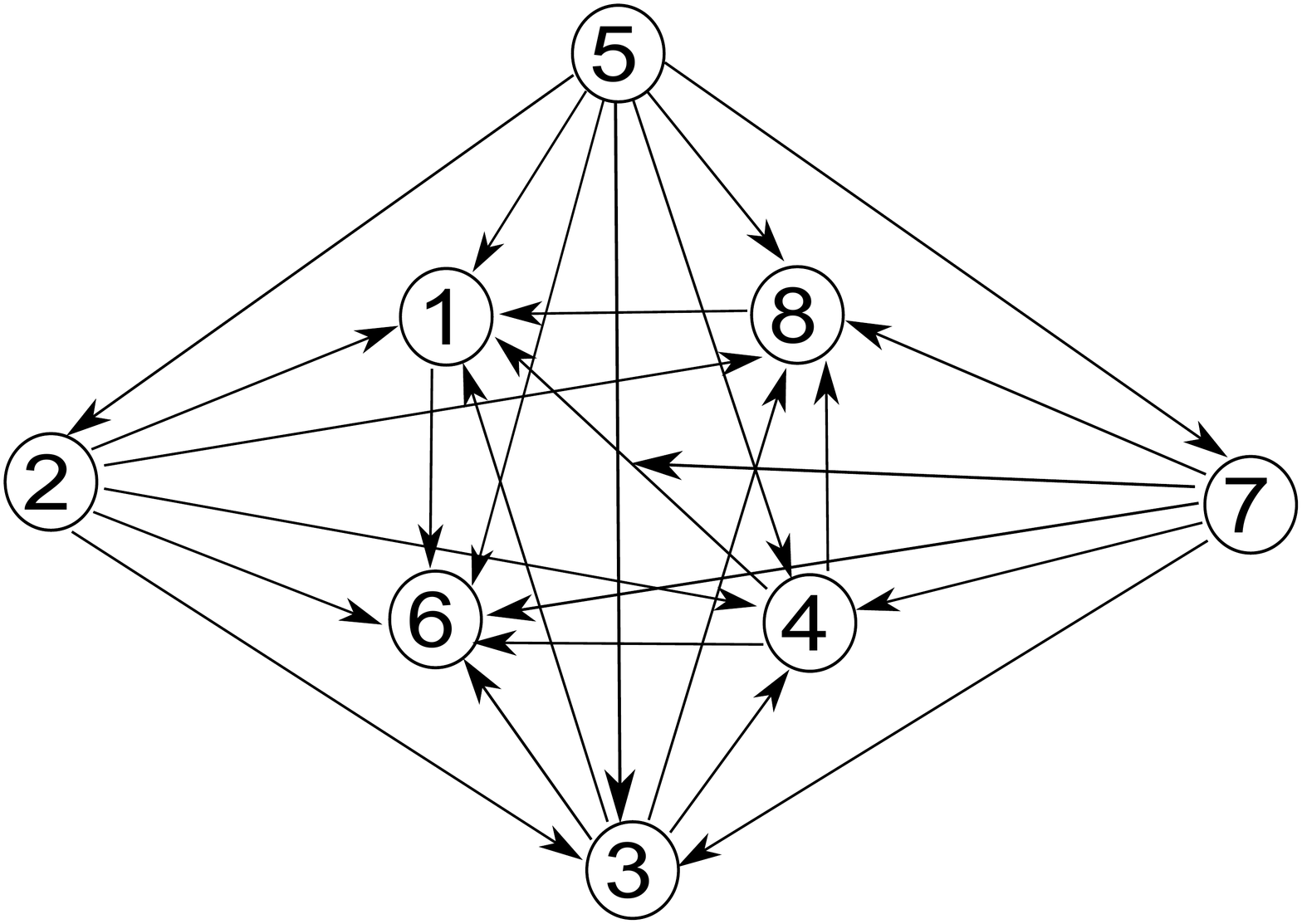}
		\label{fig:2b}}	
	\caption{DAGs for 5 and 8 nodes using CVS with NPM }
	\label{fig:2}
\end{figure}
For estimation of causal DAGs, we used i.i.d samples generated using non-Gaussian distributions. Results for 5 and 8 number of node sets are used to compare the causal DAGs formed by CVS with NPM and DirectLiNGAM. For both methods the causal DAGs are formed using the strength matrices and confounders in the DAG are added from error matrix. The generated causal DAGs are shown in Figures \ref{fig:2} and \ref{fig:3}.In Figure \ref{fig:3} (a), the constructed model using DirectLiNGAM formed a bi-directed edge from 1 and 3. The strength and errors matrices used in this model are supplied in the Supplementary material.
The performance of our method for real world data is compared with DLiNGAM. We used the regression, multivariate and categorical dataset of Boston Housing, available in UCI Repository [4]. Dataset contains 14 variables and 506 instances. For causal construction we removed the third variable from the dataset, as it only contains the binary values. The causal DAG for 13 nodes of Boston data is shown in Figure \ref{fig:4}.
As shown in Figure \ref{fig:4}, CVS with NPM always produces a DAG whether the data is noisy or non-Gaussian, so acyclic condition always holds.

\section{Conclusion}

Our method provides a new approach to causal inference and structural learning focusing to measure the strength of informations passed in the nodes. Proposed model provides solutions for how much information is passed from parent node to child and how much a child node is affected by external confounders, which were remains unsolved till date. The novelty of the method is to provide measures for path coefficient in units of percentage contributions from a rang of \(\left[0,1\right]\). The model provides strong criteria for directed acyclic causal graph formation as shown in above figures. The causal DAG estimation capability is fully verified in synthetic and real world conditions in comparison to other methods and results shown in Section \ref{sec: 5}. The method is very useful in noisy datasets to produce causal DAGs. More details about the noisy data results and reversible system is provided in supplementary material.  A complete R code is available for further utilization and development and all the dataset used is provided in the supplementary material. The results of the proposed method can be used to provide an image tool for object oriented plotting of causal DAGs, which is a future perspective of this proposal. The major problem of this future work is the arrangement of multivariate nodes for bigger models. The NPM may  prove it's usefulness in broad areas of studies and collider v-structures can be used with other Bayesian methods for structure estimations of types non-confounder and noisy systems. 




\subsubsection*{References}

\small{	
	
[1] Bach, F. R. \&  Jordan, M. I. (2002) Kernel independent component analysis. {\it Journal of Machine Learning Research}, 3:1-48.

[2] He, Yang-Bo \& Geng, Zhi. (2008) Active Learning of Causal Networks with Intervention Experiments and Optimal Designs.  {\it Journal of Machine Learning Research}, 9:2523-2547.

[3] Lauritzen, S.L. (1996) {\it Graphical Models}, Clarendon Press, Oxford.

[4] Murphy, P. M. \& Aha, D. W. (1995) {\it UCI repository of machine learning databases}. Available at \url{ http://www.ics.uci.edu/~mlearn/MLRepository.html}.

[5] Pearl, J. (2000) Causality: Models, Reasoning, and Inference. {\it Cambridge University Press}, 2nd ed. 2009.

[6] Pellet, J.P. \& Elisseeff, A. (2008) Using Markov Blanket for Causal Structure Learning. {\it Journal of Machine Learning Research}, 9:1295-1342.

[7] Shimizu, S., Hoyer, P. O., Hyv{\"a}rinen, A. \& Kerminen, A. (2006) A Linear non-Gaussian Acyclic Model for Causal Discovery. {\it Journal of Machine Learning Research}, 7:2003-2030.

[8] Shimizu, S., Inazumi, T., Sogawa, Y., Hyv{\"a}rinen, A., Kawahara, Y., Washio, T., Hoyer, P. O. \& Bollen, K. (2011) DirectLiNGAM: A Direct Method for Learning a Linear non-Gaussian Structural Equation Model. {\it Journal of Machine Learning Research}, 12:1225-1248. 

[9] Uhler, C., Raskutti, G., Buhlmann, P. \& Yu, B. (2013) Geometry of the Faithfulness Assumption in Causal Inference. {\it The Annals of Statistics},{\bf Vol.}{ 41}, No. 2, 436-463.

[10] Xie, X. \& Geng, Z. (2008) A Recursiv Method for Structural Learning of Diirected Acyclic Graphs. {\it Journal of Machine Learning Research}, 9:459-483.

}
\end{document}